\documentclass[conference,a4paper]{IEEEtran}

\ifCLASSINFOpdf
  \usepackage[pdftex]{graphicx}
  \graphicspath{{../figures/}}
  \DeclareGraphicsExtensions{.pdf,.jpeg,.png}
\fi

\usepackage{graphicx}
\usepackage{amsmath,amssymb} 
\usepackage[colorlinks]{hyperref}
\usepackage{cleveref}
\usepackage{array}
\usepackage{caption}
\usepackage{subcaption}
\usepackage{multirow}
\usepackage{algorithm}
\usepackage{algpseudocode} 
\usepackage{mathtools}
\usepackage{color}
\usepackage{array}

\newcolumntype{M}[1]{>{\centering\arraybackslash}m{#1}}

\begin{document}
\title{A Classification Methodology based on \\Subspace Graphs Learning}

\author{
\IEEEauthorblockN{Riccardo La Grassa, Ignazio Gallo, Alessandro Calefati}
\IEEEauthorblockA{University of Insubria,\\
Department of Theoretical and Applied Sciences,\\
Varese, Italy\\
Email: \{rlagrassa, ignazio.gallo, a.calefati\}@uninsubria.it
\and
\IEEEauthorblockN{Dimitri Ognibene}
\IEEEauthorblockA{University of Essex,\\
School of Computer Science and Electronic Engineering,\\
Colchester, UK \\
Email: dimitri.ognibene@essex.ac.uk 
}
}
}

\maketitle
\begin{abstract}
In this paper, we propose a design methodology for one-class classifiers using an ensemble-of-classifiers approach. 
The objective is to select the best structures created during the training phase using an ensemble of spanning trees. 
It takes the best classifier, partitioning the area near a pattern into $\gamma^{\gamma-2}$ sub-spaces and combining all possible spanning trees that can be created starting from $\gamma$ nodes.
The proposed method leverages on a supervised classification methodology and the concept of minimum distance.
We evaluate our approach on well-known benchmark datasets and results obtained demonstrate that it achieves comparable and, in many cases, state-of-the-art results.
Moreover, it obtains good performance even with unbalanced datasets.
\end{abstract}

\section{Introduction}
In Machine Learning, the usage of multiple classifiers (ensemble-of-classifiers) is a well-known technique employed to get a boost on performance compared to single classifiers~\cite{gallo2017multimodal}.
Even though it has been applied to different scenarios with great results, the sensitivity of the system on long tailed datasets still remains an open problem~\cite{blaszczynski2015}.
Another key aspect related to ensemble learning is the margin~\cite{schapire1998}: several studies have shown that the generalization ability of an ensemble of classifiers is strictly bound to the distribution of its margins on the training examples~\cite{schapire1998,tax2002using}. 
A good margin distribution means that most examples have large margins~\cite{hu2014exploiting}. 
Moreover, ensemble margin theory is an effective way to improve the performance of classification models~\cite{cantador2005}.
This theory has been successfully applied in unbalanced data sampling~\cite{fan2011,qian2014,liu2009,feng2017},
noise removal~\cite{feng2017investigation,feng2015identification,feng2015class}, instance selection~\cite{marchiori2009class}, feature selection~\cite{alshawabkeh2013hypothesis} and classifier design~\cite{gao2010kth,li2013dynamic,xie2012margin}.
In decision-boundary classifiers, finding best margin settings is a hard task because many parameters must be estimated in order to obtain a good approximation of the final model.
In recent works~\cite{livi2016,jeong2019,snell2017} many graph-based models with optimization are introduced.
These approaches combine graph theory structures and Machine Learning, modelling the data within the euclidean space.
Their aim is to find a better boundary representation in order to make more accurate predictions.
In this paper, we developed a novel approach to improve classification results.
Main contributions of our paper can be summarized as follows:
\begin{itemize}
    \item A design methodology for one-class classifiers capable of predicting the class of a pattern using different partitions created from the surrounding area.
    \item An objective function based on information extracted during the training phase which selects the best partition among all possible $\gamma^{\gamma-2}$ combinations, built on top of most similar objects from a specific target.
\end{itemize}

\section{Related Work}
In literature, there are plenty of works dealing with decision-boundary models.
For instance, Pekalska~\textit{et al.}~\cite{pekalska} use a non-parametric model based on one-class classifiers with a decision-boundary handled through a threshold.
In order to handle margins in the final model a threshold based on median (or a parameter) of weighted-edge of minimum spanning tree is set to expand or deflate the decision-boundary.
They achieve good results also on long-tail datasets.
However, the high computational training time required to find the minimum spanning tree in a high-dimensional dataset represents a drawback.
In~\cite{livi2016} authors describe an interesting graph-based approach for one-class classifiers with optimizator learning in a binary classification problem finding the best partition exploiting a criterion based on mutual information minimization.
The idea is original and uses many complex concepts based on graph theory and subspace learning. 
They don't use an ensembles approach to find out best partitions but provide a non-parametric model to estimate information-theoretic quantities, such as entropy and divergence. 
The decision regions are computed partitioning the k-nearest neighbour graph vertices related to the components of the training set. 
The main issue of this approach is the high computational time required caused by the computation of all possible combinations.

Many recent state-of-the-art works employ a distance metric for classification. 
Snell~\textit{et al.}~\cite{snell2017} compares the euclidean distance between the query example and the mean class of embedded support examples. 
Authors propose this prototypical network for pattern recognition, where a classifier, given few examples of training, creates a model able to achieve excellent results.
However, a weakness of this approach is about possible outliers within a class that make the mean unstable and the problem of the overlapping patterns among classes. 
Furthermore, considering only few examples or using $k$ object to search the neighbours of an object can lead to bad partition of that area that result in wrong classifications.

In~\cite{kabir2018} authors propose a mixed bagging model where each bootstrap have varying degrees of hardness. 
Instance hardness is the probability that an instance will be misclassified by a classifier built from other instances of the dataset. 
Outliers, instance near the decision boundary and instances surrounded by other instances from the opposite class are usually harder to classify \cite{smith2014instance}, they propose to remove the harder instances from the training set to obtain a clear model, instead to use random bootstrap created through a bagging method, they~\cite{kabir2018} create different subsets considering the hardness of a group of instances. 
Our work, use a generator of spanning trees (we calculate all possible spanning trees) near to the neighbours of an object to classify, and create $\gamma^{\gamma-2}$ initial partitions. 
The selection criterion to reduce these subsets is based on bagging approach and will be widely discussed into proposal approach. 
For fair comparison, we use the same $47$ publicly available datasets for binary classification problem, used in~\cite{jeong2019,kabir2018}.


In~\cite{LaGrassa}, La Grassa~\textit{et al.}, creates a binary model based on the combination of two one-class classifiers. 
Authors propose three non-parametric approaches that improve the model described in~\cite{pekalska} showing good results in term of final accuracy on many well-known datasets.
More precisely, the first model creates two minimum spanning trees, one per class, using weighted edges based on the euclidean metric. 
However, strongly nonlinear datasets with unbalanced or high dimensional data can be a serious problem in terms of accuracy or performance.
In order to improve this issues,~\cite{LaGrassa} creates two models based on a combination of two minimum spanning trees (or n-ary trees), created from the neighbourhood of an instance to classify, instead of the whole training set, avoiding data overlapping from different classes and partition different areas to make the classification. 
In this way, authors, show better results in terms of accuracy and computational time, also investigating the size of MST created by neighbours. 
The main idea is very interesting but suffers from an issue: the rule to adjust the boundary decision from neighbours is always the same (MST) and that model does not discover new structures to improve the performance, furthermore, as a non-parametric models, they do not extract parameters from the training phase.

Major differences of our approach from existing ones are listed below:
\begin{itemize}
\item We do not create a close partition to make predictions, but we use jointly a spanning tree and a threshold.
This information is useful to extract some parameters during the training phase.

\item We use the information coming from the training phase to search similar partitions around an instance to classify and combine decision boundaries to make final predictions based on a ensemble method.

\item We use euclidean metric instead to use entropic information by spanning graph.

\item We compute all spanning trees starting to k-neighbours of an instance $x$ to find partitions.
\end{itemize}


\section{The proposed approach}
\begin{figure}[ht]
\centering
    \includegraphics[scale=.45]{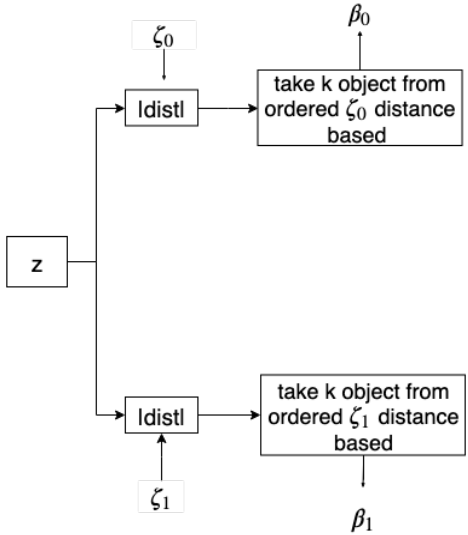}\par 
    \caption{Pre-test step: The model computes all euclidean distance between the instance to classify and other instances of the datasets. Then, it takes the information of instances nearer to an instance $z$. 
    Finally, we extract beta value, using different parameters described in~\ref{parameters_list}.}
    \label{fig:spt_beta}
\end{figure}

In this work, we introduce a model that extracts parameters from the training phase that will be used during the test in order to increase the performance in terms of final accuracy. 
The strength of our model is that it creates $\gamma^{\gamma-2}$ different sub-spaces creating all possible spanning trees, as described in~\cite{read1975bounds}, and selects the best structures that will be used during the test, as shown in Fig.~\ref{fig:spt_beta}. 

\subsection{Partition of the dataset}
The classifier used in this work is the same used in~\cite{LaGrassa}, but we partition different area to try to discover a new pattern able to recognize correctly new instances.


We consider a binary classification task. 
Given a training set $X=\{x_{0}, x_{1},\dots,x_{n-1}\}$, where a generic $x_{i}$ is a vector defined as: 
\begin{equation}\label{instance}
 x =[a_{0},a_{1},\dots,a_{m-1}] 
\end{equation}
 $X$ is  partitioned in two subsets $S$ and $(X-S)$. 
 In turn $(X-S)$ is partitioned in $X_{0}\bigcup X_{1}$ which respectively contain all the elements with label $1$ and  $-1$. 
%
%
Here, $|X|=n$ and $|x|=m$ are respectively the size of training set and a generic instance $x \in X$.

\subsection{Creation of Spanning Trees}
We use a discriminative function~\cite{LaGrassa} (modified under certain condition, see algorithm  \ref{alg:spt_cd test}) $f_{1}({x)}\xrightarrow{}y$ capable of mapping a generic vector $x$ into a label $y$. 
For each $s\in S$ and $X_{i}$ with $i \in \{0,1\}$, we compute the  euclidean distance and create the lists of pairs:
\begin{equation}
\begin{aligned}
\label{euclidean_distance_x0}
D_{i}(s)=\{(x_{j},\sqrt{(s - x_{j})^2}) \mid x_{j} \in X_{i}, s \in S, s \neq x_{j} \}\\
\end{aligned}
\end{equation}

The lists $D_{i}(s)$ are sorted in ascending order of distance, the second element of the pair.
For each element $s$ of the set, we take the $\gamma$ nearest elements, formally:
\begin{equation}\label{setV}
\begin{aligned}
V_{i}(s)=\{d_{j,1}\ \mid d_{j} \in D_{i}(s), 0 \le j \le \gamma \}\\
\end{aligned}
\end{equation}

Starting from $V_{0}(s)$ and $V_{1}(s)$, we build the two corresponding sets of spanning trees $H_{0}(s)$ and $H_{1}(s)$.

\begin{equation}\label{H0,H1}
\begin{aligned}
H_0(s)=\{h_{0},h_{1},\dots,h_{(\gamma^{\gamma-2})-1}\}\\
H_1(s)=\{h'_{0},h'_{1},\dots,h'_{(\gamma^{\gamma-2})-1}\}
\end{aligned}
\end{equation}

where $h=(V,E)$ is a generic spanning tree, $V\in \{V_0(s),V_1(s)\}$ the sets of nodes of its $\gamma$ nearest neighbours and $E$ the selected set edge.
For each $s \in S$, we take all pairs of spanning trees in $H_{0}(s)$ and $H_{1}(s)$ and use the classifier described in~\cite{LaGrassa} using a Bootstrap Aggregating approach to select a subset of them (SPT\_CD) for each label class (see Fig.~\ref{fig:train_image}).

\subsection{Selection of best sub-graphs}
We define two sets $\zeta_0$ and $\zeta_1$ that represent two dictionaries used in the test phase to predict the labels.
Each key of this dictionary is an element of a class of the training set and values are all spanning trees that performed a correct classification using the classifier proposed in~\cite{LaGrassa}.
For each element, we store also the euclidean distance from the spanning tree and the weighted sum of all edges. 
Finally, the dictionary will contain a set of triplets for each instance based on right classifications. 
More formally:

\begin{equation}
\begin{aligned}
    \zeta_{0}=\{(s,q)\mid s \in S, q = \Gamma(s,h,H_{1}(s))\  \forall h \in H_{0}(s) \}\\
    \zeta_{1}=\{(s,q)\mid s \in S, q = \Gamma(s,h,H_{0}(s))\  \forall h \in H_{1}(s) \}
\end{aligned}
\end{equation}

where $\Gamma$ is the function $\Gamma : s \rightarrow q$:
\[ \Gamma(s,h,H_{j}(s)) =
\begin{cases}
(h, d(s,h), \sum_{e \in h}w(e))  & \quad \text{if } C \ge 0 \\
\emptyset & \text{otherwise}
\end{cases}
\]
$h$ is a specific spanning tree selected from all possible generated, $d$ is the distance between $s$ to $h$ and the last element of the triplet represents the weighted sum of every couple of spanning trees.

We use a counter $C$ to assign a label:
\begin{equation}\label{bagging counter}
 C =
\begin{cases}
+1 & \text{if SPT\_CD}(H_{j}(s),h) = y\\
-1 & \text{otherwise}
\end{cases}
\end{equation}

where $y$ is the label of s.

We sort $\zeta$ in ascending way using the already computed  distances $d(s,h)$.
Finally, we use these dictionaries for the classification task described in the next section.

\begin{figure}[ht]
\centering
    \includegraphics[scale=.16]{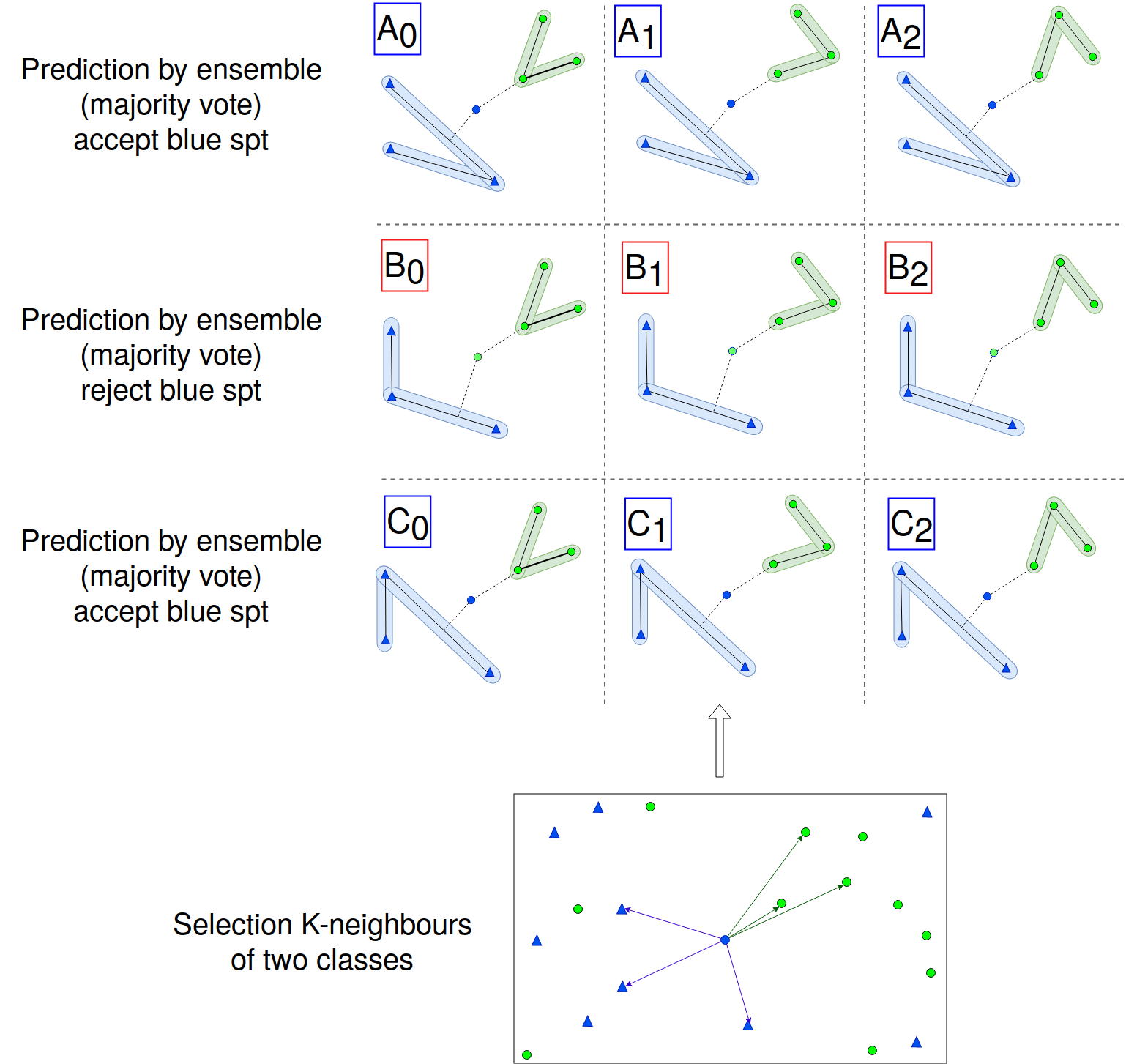}\par 
    \caption{Ensemble methodology used in the training phase to select spanning trees that correctly classify an instance $x$. 
    The information of these spanning trees will be combined and used in an objective function to improve the accuracy in the test phase. 
    Blue points represent instances from class $1$ and green points represents class $-1$. 
    In this example, we fix $\gamma=3$ to create all spanning trees.}
    \label{fig:train_image}
\end{figure}

In Fig.~\ref{fig:train_image}, we show a toy example of the training phase, based on graphs created considering k-neighbours of a specific object (see Eq.~\ref{H0,H1}). 
We highlight the ensemble approach applied in each row $A,B,C$ where a spanning tree recognizes the pattern, while the other refuses it. 
If the number of positive responses is greater than the number of rejections, we keep that spanning tree. 
As described before, we are partitioning the neighbourhood of an element to get a model (decision-boundary) able to make the right predictions. 
For instance, in the first row the blue object is correctly classified by the model generated with same label, but in the second row it is refused leading to wrong prediction (see pseudocode~\ref{alg:training_pseudocode}).
\subsection{Classification Task}
In test phase, we use Eq.~\ref{euclidean_distance_x0} and define a vector $S^*$ as the set of distances between a generic instance of test $z$ and each $x \in X$. 
For easier understanding, we introduce the pseudocode~\ref{alg:beta_assignment}.

\begin{algorithm} \caption{Training step}
\label{alg:training_pseudocode}
\scriptsize
\begin{algorithmic}[1]

\For {$s \in \mathcal S $}
    \For {$x \in \mathcal X_{0}$} \textit{class 1}
        \State  $euclidean\ distance \ list \gets (|| s - x ||, x)$
\State Sort euclidean distant list    
\State Take k-neighbours of s and create a complete graph with them
\State Generate all spanning trees 
\EndFor
\EndFor

\For {$s \in \mathcal S $}
    \For {$x \in \mathcal X_{1}$} \textit{class -1}
        \State  $euclidean\ distance\ list= \gets (|| s - x ||, x)$
\State Sort euclidean distant list    
\State Take k-neighbours of s and create a complete graph with them
\State Generate all spanning trees
\EndFor
\EndFor

\For {$h \in \mathcal H_{0}(s) $}
    \For {$h' \in \mathcal H_{1}(s)$}
    \State prediction SPT\_CD(h, h') (see eq. \ref{bagging counter})
\EndFor  

\State {majority vote ($C \ge 0$)}
\State {$dictionary_{0}$ $\gets$ Save (h, distances (s, h), sum\_weighted h)}
\EndFor

\State \textbf{repeat line 17-21 for each} $h' \in H_{1}(s) \ and \ h \in H_{0}(s)$ and 
\State $dictionary_{1}$ $\gets$ Save (h', distances (s, h'), sum\_weighted h')

\end{algorithmic}
\end{algorithm}

\begin{algorithm} \caption{Beta assignment (pre-test step)}
\label{alg:beta_assignment}
\scriptsize
\begin{algorithmic}[1]
\State {\textit{Given Z be the test set}}
\For {$z \in \mathcal Z $}
    \For {$x \in \mathcal X_{0} $}
    \State  $S^* \gets (|| z - x ||, x)$
    \EndFor
\EndFor

\State {sorted(euclidean\_distance)}

\For {$i \in \mathcal [1,k] $}
\State {best\_similarity= [$r[3]$  for r in $\zeta_{0}$($S^* [i][2])]$}
\EndFor

\State {$weight\_all\_spt$ = sorted($best\_similarity$)}

\State{$\theta_{0}$ = $|weight\_all\_spt| * alpha\_best\_sub\_graph$}

\State{$\beta_{0} = weight\_all\_spt_{\theta_{0}}$}
\State {\textbf{repeat line 2-13 considering for each $x \in X_{1}$, $\zeta_{1}$, $\theta_{1}$} and find $\beta_{1}$}
\State return $\beta_{0}, \beta_{1}$
\end{algorithmic}
\end{algorithm}

\begin{algorithm} \caption{Classify($z$)}
\label{alg:test_pseudocode}
\scriptsize
\begin{algorithmic}[1]
\State $\beta_{0}, \beta_{1}$ = Call Algorithm \ref{alg:beta_assignment}: \textit{Beta assignment} 
\State N0=getK-Neighbours(z,X0,k)
\State T0=getSPTs(z,N0)
\State N1=getK-Neighbours(z,X1,k)
\State T1=getSPTs(z,N1)

\State $H_{0}^*$,$H_{1}^*$ = Sub-graphs selection(T0,T1) (see Equ~\ref{eq:gradient})

\For {$h \in \mathcal H_{0}^*(s) $}
    \For {$h' \in \mathcal H_{1}^*(s)$}
    \State{\textbf{SPT\_CD(h, h')}\ with couple of spt's  returning a prediction $\pm 1$} see eq. \ref{bagging counter}
\EndFor  

\State majority vote ($C \ge 0$) for each $h \in H_{0}(s)$
\EndFor
\State Final majority vote ($C\ge 0$) and assign prediction
\end{algorithmic}
\end{algorithm}

\begin{algorithm} \caption{SPT\_CD (h, h', z)}
\label{alg:spt_cd test}
\scriptsize
\begin{algorithmic}[1]
\For {$(x_{i},x_{j}) \in E(h)$ }
\If {$0<=\frac{(x_j - x_i)^T * (z-x_i)}{||x_j - x_i||^2}<=1$}
\State {$P_{e_{_{i_{j}}}}(z)=x_{i} + \frac{(x_{j} - x_{i})^T * (z-x_i)}{||x_j - x_i||^2}*(x_j-x_i)$}
\State {$d(z|e_{_{i_{j}}}) \gets ||z - P_{e_{_{i_{j}}}}(z)||$}
\Else
\State{$d(z|e_{_{i_{j}}}) \gets min \big\{ ||z - x_i||, ||z -x_j||\big\}$}
\EndIf
\EndFor
\State \textbf{Repeat line 1-8 for spanning tree} $h'\ of\ H_{1}$\\
{\textit{Find the boundary of spanning tree h of $H_0$ and h' of $H_1$ based on $\alpha$ parameter}
\State $e(h) = (||e_0||,||e_1||,..||e_n||)$
\State $e(h') = (||e_0||,||e_1||,..||e_n||)$
\State $\theta_0= || e_{(\alpha n )}||$
\State $\theta_1= || e_{(\alpha n )}||$

\State min dist0 = $min(d(z|e_{ij}))$
\State min dist1 = $min(d1(z|e_{ij}))$\\
\textit{make prediction}}
\State 1 $\gets$  \text{if  $d_{SPT\_CD_0}(z|h) <= \theta$ and $d_{SPT\_CD_1}(z|h) > \theta_1$} 
\State -1 $\gets$  \text{if  $d_{SPT\_CD_0}(z|h) > \theta$ and $d_{SPT\_CD_1}(z|h) <= \theta_1$}
\If{\textbf{min dist0 $<= \theta$ and min dist1 $<= \theta_1$}} 
\State knn weight1=order($d_{1}(z|u)$) and take $k_1$-elements 
\State knn weight0=order($d_{0}(z|v)$) and take $k_1$-elements
\State euclidean distance vectors = (knn weight1 - knn weight0)
\State positive $=$ Count $n_i > 0$ in euclidean distance vectors
\State negative $=$ Count $n_i < 0$ in euclidean distance vectors
\If{negative $>=$ positive}
\State prediction $\gets 1$
\Else
\State prediction $\gets -1$
\EndIf
\EndIf
\If{\textbf{min dist0 $> \theta$ and min dist1 $> \theta_1$}}
\State The approach is equal to lines [20-31]
\EndIf
\State \textbf{return}
\end{algorithmic}
\end{algorithm}

Furthermore, in Fig~\ref{fig:train_image} we consider a boundary decision derived from the weighted edge sum of a specific spanning trees (blue and green boundaries).  
The reason we don't consider only the edge couples among k-neighbours of instance nearest to it (in training and test step) is to create a boundary more detailed thanks to his neighbourhood. 
This task plays a crucial role in the classification task because is able to model instances of the same classes and to represents these as a unique decision boundary.
Similarly, we apply the Pseudocode~\ref{alg:beta_assignment} considering the euclidean distance between $z$ and each $x\in X_{1}$ and $\zeta_{1}$
Then, we find all spanning trees considering the neighborhood of $z$ using Eq.~\ref{euclidean_distance_x0} and Eq.~\ref{setV} to find Eq.~\ref{H0,H1}.
Finally, we choose two subsets of spanning trees $\widetilde{H}_{0} \subseteq H_{0}$ of $z$ and $\widetilde{H}_{1}\subseteq H_{1}$ of $z$  such that:
\begin{equation}\label{eq:gradient}
\Delta \widetilde{H}_{z}(\beta ,h)= \sum_{e \in h}{w(e) - \beta }
\end{equation}
where $w(e)$ is the weight for an edge of the considered spanning tree and $\beta$ is the weighted sum of a spanning trees extracted considering a index from $weighted\_all\_spt$ (see~\ref{alg:beta_assignment}).
We use a parameter $best\_spt$ to consider only a subset of distance-ordered $\Delta \widetilde{H}_{x}$. 
We compute all minimum distances between an object with a specific sub-graph according to the orthogonal projection of $x$ onto a line of sub-graph or minimum Euclidean distance.
Given a training set $X$, split into two sets $X_{0},X_{1}$, we consider $\widetilde{H}_{0}$ and $\widetilde{H}_{1}$ as sets of spanning trees. 


From the previously computed $H^*_{0}$, $H^*_{1}$ sets, we create all possible combinations and sum the number of correct classifications for a pattern $z$.
The label assigned is the one which obtained most votes according to the bagging method and using SPT\_CD as classifier.
In other words, we define a learning approach to find the best partitions through the $k$ nearest neighbours of the target object.
Let us define the objective function:
$$\eta(h,H(s)) = \frac{1}{1+\Delta \widetilde{H}_{x}(\beta ,h)}$$
that goes to $1$ when the variation is low and goes to $0$ when the variability of data becomes higher.
Therefore, the problem is to minimize the following function:
$$\min_{h,H(s)}\eta(h,H(s))$$
Therefore, in the prediction phase, we use the same the classifier of \cite{LaGrassa} under certain conditions (see Pseudocodes \ref{alg:beta_assignment}, \ref{alg:test_pseudocode}, \ref{alg:spt_cd test}) considers two aspects:

\begin{itemize}
    \item Projection of point $x$ on a line defined by vertices ${x_i, x_j}$
    \item Minimum Euclidean distance between $(x, x_i)$ and $(x, x_j)$
\end{itemize}

The projection of $x$ is defined as follow:
$$p_{e_{i,j}}(x)=x_i + \frac{(x_j - x_i)^T(x - x_i)}{||x_j - x_i||^2}(x_j - x_i)$$
if $p_{e_{i,j}}(x)$ lies on the edge $e_{i,j}$= ($x_i$,$x_j$), we compute $p_{e_{i,j}}(x)$ and the Euclidean distance between $x$ and $p_{e_{i,j}}(x)$, more formally:
$$0<=\frac{(x_j - x_i)^T(x - x_i)}{||x_j - x_i||^2}<=1$$ then
$$d(x|e_{i,j})=||x - p_{e_{i,j}}(x)||$$
Otherwise we compute the Euclidean distance of $x$ and pairs ($x_i$, $x_j$), precisely:
$$d(x|e_{i,j})=min(||x - x_j||,||x - x_i||)$$

Then, a new instance $x$ is recognized from a spanning tree if it lies within the boundary, otherwise, the object is considered as outlier.
The decision whether an object is recognized by classifier or not is based on the threshold of the shape created in this phase, more formally:
$$d_{h}(x|X) <= \theta$$
The threshold $\theta$ is a parameter used to assign the boundary dimension of a spanning tree. 
Then, given $\hat{e}=(||e_1||,||e_2||,...,||e_n||)$ as an ordered edge weights values, we define $\theta$ as $\theta = ||e_{[\alpha n]}||$, where $\alpha \in [0,1]$. 
For instance, with $\alpha=0.5$, we assign the median value of all edge weights of the spanning tree.

\begin{figure*}
\centering
    \includegraphics[scale=.45]{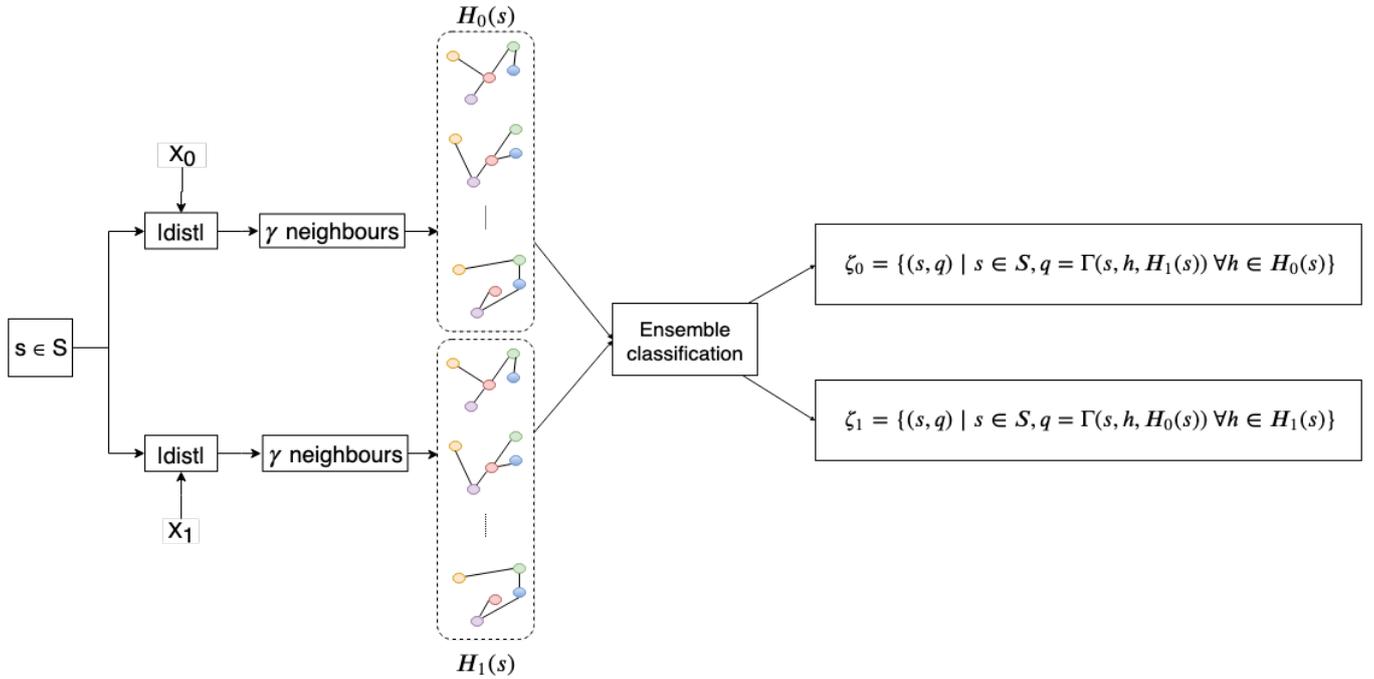}\par 
    \caption{Training step: For each instance in the training set, the model uses Euclidean distance than other instances into datasets split per classes. The model generates all possible spanning tree and applies an ensemble method based on majority vote using the same classifier in \cite{LaGrassa} on different decision boundaries. Finally, the model saves the best-spanning trees, weighted sum of them and euclidean distance between instance and the spanning trees.}
    \label{fig:spt_schema}
\end{figure*}
\begin{figure*}
\centering
    \includegraphics[scale=.5]{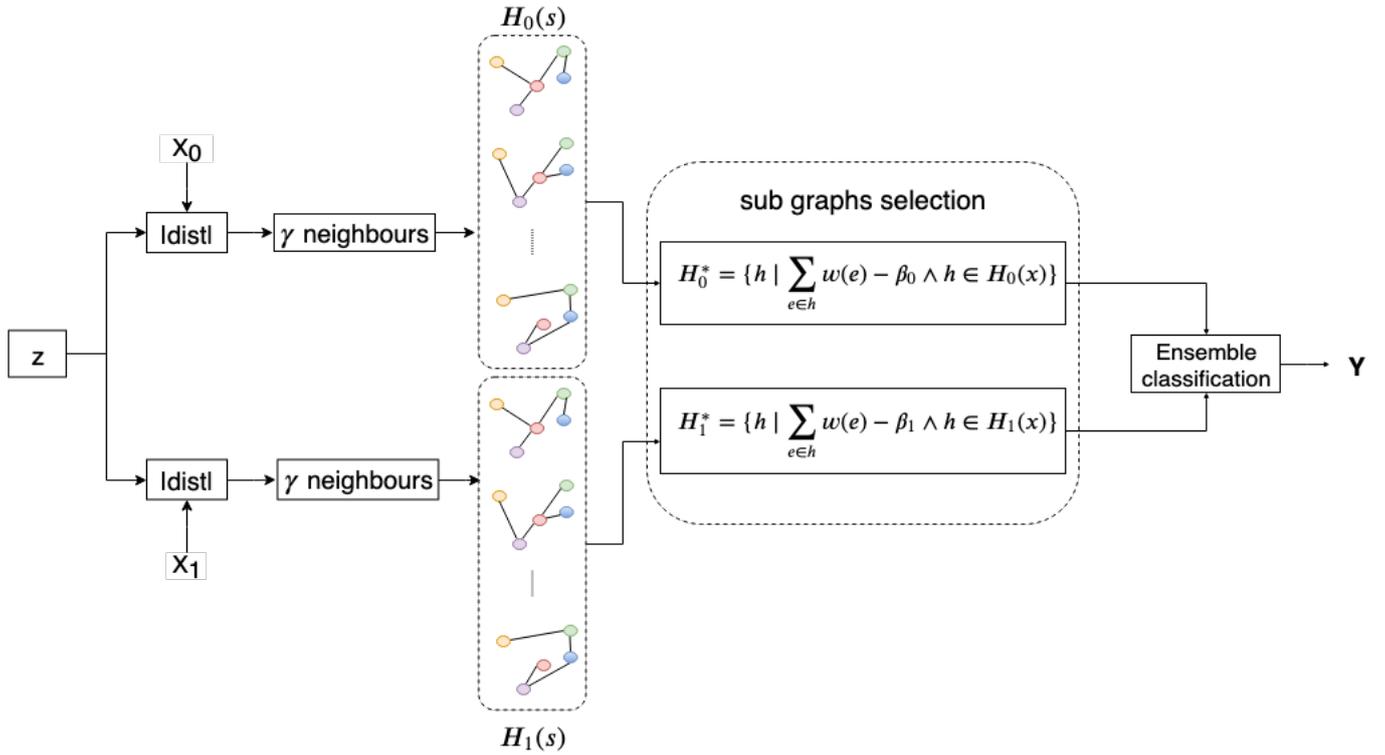}\par 
    \caption{Test step: For each new instance, the model computes all euclidean distance with train split per classes and generate all possible spanning trees from his neighbourhood.
    Then, we use the parameters extracted in the training step to minimize an objective function to select a subset of spanning trees generated. Last, we apply an ensemble method based on majority vote to classify the instance.}
    \label{fig:spt_test}
\end{figure*}

\section{Datasets}
To evaluate the effectiveness of our approach, we tested it on many well-known datasets.
As can be seen from Table~\ref{tab:datasets}, there is huge variability in terms of the number of features for each dataset.
Another aspect which makes experiments more challenging is the displacement of patterns between classes.

\begin{table}
\centering
\caption{Number of features, instances and positive-negative samples for all the datasets used in our experiment.}
\label{tab:datasets}
\begin{tabular}{l|ccc} 
\hline
Datasets (Acronyms)                 & Features      & Instances     & pos-neg  \\ \hline
Arcene (A)                          & 10000         & 100           & 44-56  \\ \hline
Gisette (G)                         & 5000          & 6000          & 3000-3000  \\ \hline
Hill (HL)                           & 101           & 606           & 305-301 \\ \hline
Sonar (S)                           & 60            & 208           & 97-111 \\ \hline
Pima (PM)                           & 8             & 768           & 268-500 \\ \hline
Banana (BN)                         & 2             & 5300          & 2924-2376 \\ \hline
Banknote authentication (BK)        & 4             & 1372          & 762-610 \\ \hline
Breast cancer ws Pr (BC-P)          & 32            & 198           & 151-47 \\ \hline
Liver (L)                           & 6             & 345           & 145-200 \\ \hline
Breast cancer ws (BW)               & 9             & 699           & 458-241 \\ \hline
\end{tabular}
\end{table}

\section{Experiments}
\begin{figure*}
\centering
    \begin{subfigure}{0.5\linewidth}
\includegraphics[width=\linewidth]{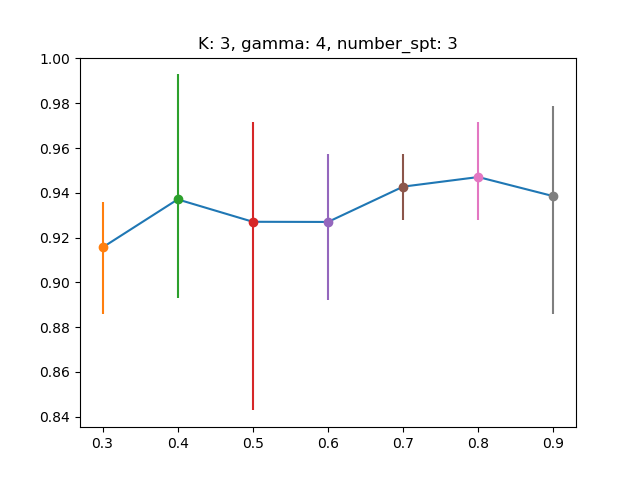} 
\label{fig:1a}
    \end{subfigure}\hfill
    \begin{subfigure}{0.5\linewidth}
\includegraphics[width=\linewidth]{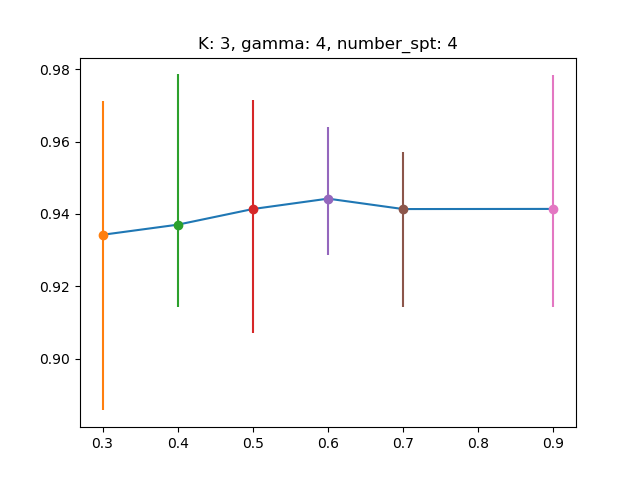}
\label{fig:1b}
    \end{subfigure}
    \caption{Average accuracy (dot) on 5-fold cross-validation on different train size of the Breast cancer Wisconsin datasets. Vertical lines represent the upper (max accuracy) and lower (min accuracy) bound of 5-fold.}

    \begin{subfigure}{0.5\linewidth}
\includegraphics[width=\linewidth]{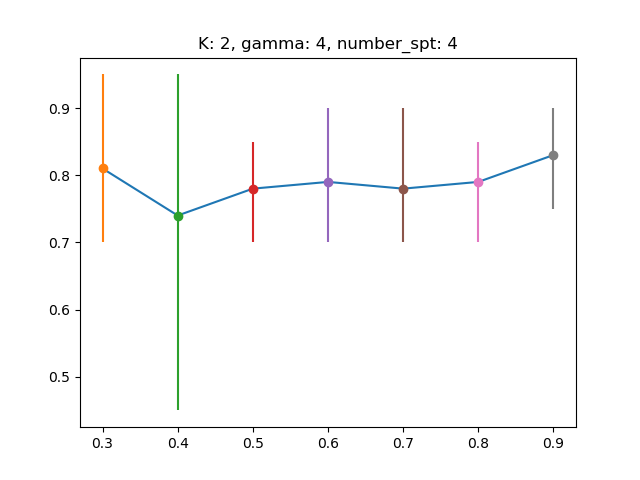}
\label{fig:2a}
    \end{subfigure}\hfill
    \begin{subfigure}{0.5\linewidth}
\includegraphics[width=\linewidth]{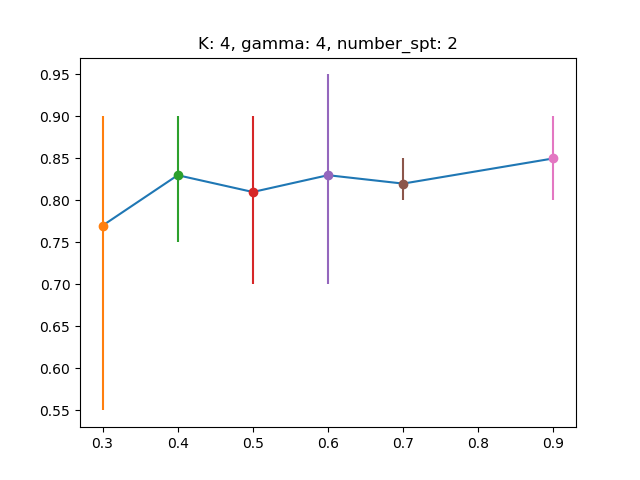}
\label{fig:2b}
    \end{subfigure}
\caption{Average accuracy (dot) on 5-fold cross-validation on different train size of the Arcene datasets. Vertical lines represent the upper (max accuracy) and lower (min accuracy) bound of 5-fold.}
    \label{fig:variation_training_set}
    
\end{figure*}

Performance of $SPT\_CD$ is evaluated using well-known metrics: Sensitivity, Specificity, Precision, F1 score, AUC score and accuracy.
In addition to these measures, we computed the ROC curve. 
In our experiments, we employed a 5-Fold Cross Validation approach and computed the average testing on different parameters described in~\ref{parameters_list}.
The ratio between train and test sets is 80\% and 20 \% respectively.
For a deeper evaluation of the proposed approach, we used also different ratios, from 30\% to 90\% of the training split~\ref{fig:variation_training_set}, to measure the impact of the training size on the speed of training and on the final accuracy. 
In all our experiments we have not applied any feature selections or deep feature extraction models on datasets.
In Table~\ref{tb3} we compare our model in terms of AUC score and many classifiers well-known into the literature on $5$ different datasets.
Results are comparable with the state-of-the-art and, in some cases, they outperform it.
Furthermore, we compares different metrics such as sensitivity, specificity, precision and final accuracy on three variations of Multi-layer Perceptron (Table~\ref{tab:iqbal}). 
We highlight the best performance of our model than neural networks reported. 
Tables~\ref{tab:krakovna} and \ref{tab:abpeykar} show more comparisons in terms of final accuracy, reporting some well-known classifiers and ensemble methods.

\begin{table}[]
\centering
\caption{Parameters list}
\label{parameters_list}
\begin{tabular}{llllll}
\cline{1-2}
Alpha         & Threshold used to find beta value in $weight\_all\_spt$   
                    &  &  &  &  \\ \cline{1-2}
Gamma         & Number of neighbor nodes                                                            &  &  &  &  \\ \cline{1-2}
Best\_spt     & Number of best spt to consider (in test phase)                               &  &  &  &  \\ \cline{1-2}
k\_neighbours & Number of instances (into training set) nearest to z                             &  &  &  &  \\ \cline{1-2}
K             & Number of neighborhood nodes to consider when \\ 
              & all classifiers refuse or accept z 
              &  &  &  &  \\ \cline{1-2}
\end{tabular}
\end{table}

\begin{table}[]
\caption{AUC of suggested classifiers and other one-class classifiers in \cite{livi2016}\cite{duin}.
In the first row average results by our model on 5-fold cross validation. The highest score in each row is marked in Bold}
\label{tb3}
\begin{tabular}{c|cccccc|}
\hline
\multicolumn{1}{|c|}{}                                                     & Classifiers                                 & BC-P                  & BW        & L               & S                   & PM        \\ \hline
\multicolumn{1}{|c|}{Our work}                        & \multicolumn{1}{c|}{$\textbf{SPT\_CD}$} & \textbf{0.646}        & 0.951                 & \textbf{0.655}  & \textbf{0.841}      &  0.666         \\ \cline{2-7}

\multicolumn{1}{|c|}{\multirow{3}{*}{$\cite{livi2016}$}}  & \multicolumn{1}{c|}{EOCC-MI}        & 0.569                 & 0.989     & 0.481  &           -                  &  -         \\ \cline{2-7}
\multicolumn{1}{|c|}{}                                & \multicolumn{1}{c|}{EOCC-1}             & 0.554                 & \textbf{0.99}      & 0.461  &           -                  &  -         \\ \cline{2-7}
\multicolumn{1}{|c|}{}                                & \multicolumn{1}{c|}{EOCC-2}             & 0.585                 & 0.989     & 0.536  &           -                  &  -         \\ \cline{1-7}
\multicolumn{1}{|c|}{\multirow{16}{*}{$\cite{duin}$}} & \multicolumn{1}{c|}{Gauss}              & 0.591                 & 0.823     & 0.586  &          0.603               &  \textbf{0.705}         \\ \cline{2-7}
\multicolumn{1}{|c|}{}                                & \multicolumn{1}{c|}{MoG}                & 0.511                 & 0.785     & 0.607  &          0.663               &  0.674         \\ \cline{2-7}
\multicolumn{1}{|c|}{}                                & \multicolumn{1}{c|}{Näıve Parzen}       & 0.538                 & 0.965     & 0.614  &          0.67                &  0.679         \\ \cline{2-7}
\multicolumn{1}{|c|}{}                                & \multicolumn{1}{c|}{Parzen}             & 0.586                 & 0.723     & 0.59   &          0.681               &  0.676         \\ \cline{2-7}
\multicolumn{1}{|c|}{}                                & \multicolumn{1}{c|}{k-Means}            & 0.536                 & 0.846     & 0.578  &          0.56                &  0.659         \\ \cline{2-7}
\multicolumn{1}{|c|}{}                                & \multicolumn{1}{c|}{1-NN}               & 0.595                 & 0.694     & 0.59   &          0.682               &  0.667         \\ \cline{2-7}
\multicolumn{1}{|c|}{}                                & \multicolumn{1}{c|}{k-NN}               & 0.595                 & 0.694     & 0.59   &          0.682               &  0.667         \\ \cline{2-7}
\multicolumn{1}{|c|}{}                                & \multicolumn{1}{c|}{Auto-encoder}       & 0.548                 & 0.384     & 0.564  &          0.611               &  0.598         \\ \cline{2-7}
\multicolumn{1}{|c|}{}                                & \multicolumn{1}{c|}{PCA}                & 0.574                 & 0.303     & 0.549  &          0.564               &  0.587         \\ \cline{2-7}
\multicolumn{1}{|c|}{}                                & \multicolumn{1}{c|}{SOM}                & 0.523                 & 0.79      & 0.596  &          0.655               &  0.692         \\ \cline{2-7}
\multicolumn{1}{|c|}{}                                & \multicolumn{1}{c|}{MST\_CD}            & 0.611                 & 0.765     & 0.58   &          0.671               &  0.659         \\ \cline{2-7}
\multicolumn{1}{|c|}{}                                & \multicolumn{1}{c|}{k-Centres}          & 0.584                 & 0.715     & 0.537  &          0.6                 &  0.606         \\ \cline{2-7}
\multicolumn{1}{|c|}{}                                & \multicolumn{1}{c|}{SVDD}               & 0.498                 & 0.7       & 0.47   &          0.589               &  0.577         \\ \cline{2-7}
\multicolumn{1}{|c|}{}                                & \multicolumn{1}{c|}{MPM}                & 0.053                 & 0.694     & 0.587  &          0.59                &  0.656         \\ \cline{2-7}
\multicolumn{1}{|c|}{}                                & \multicolumn{1}{c|}{LPDD}               & 0.539                 & 0.8       & 0.564  &          0.639               &  0.668         \\ \hline
\end{tabular}
\end{table}

\begin{table}[]
\caption{Comparison in terms of sensitivity, specificity, precision and accuracy between our model (SPT\_CD) and results of three variations of Multilayer Perceptron in $\cite{iqbal2018functional}$.
Acronyms in this Table are defined in~\ref{tab:datasets}.}
\label{tab:iqbal}
\centering
\begin{tabular}{c|ccccc|}
\hline
\multicolumn{1}{|c|}{Datasets}             & Method & Sens                    & Spec                      & Pr                        & Acc                           \\ \hline
\multicolumn{1}{|c|}{\multirow{4}{*}{BN}} & \textbf{SPT\_CD}    &  \textbf{85.29}                      &  \textbf{85.29}                        &  \textbf{85.1}                       & \textbf{85.3}        \\ \cline{2-6} 
\multicolumn{1}{|c|}{}                    & CMLP       & 71.97                   & 71.97                     & 80.09                     & 65.55    \\ \cline{2-6} 
\multicolumn{1}{|c|}{}                    & SGMLP      & 72.54                   & 72.54                     & 80.71                     & 69.65    \\ \cline{2-6} 
\multicolumn{1}{|c|}{}                    & CWMLP      & 77.08                   & 77.08                     & 78.95                     & 73.42    \\ \hline
\multicolumn{1}{|c|}{\multirow{4}{*}{BW}} & \textbf{SPT\_CD}    & \textbf{95.16}          & \textbf{95.16}             & \textbf{94.46}            & \textbf{95.27} \\ \cline{2-6} 
\multicolumn{1}{|c|}{}                    & CMLP       & 74.88                   & 72.13                     & 70.93                     & 78.11    \\ \cline{2-6} 
\multicolumn{1}{|c|}{}                    & SGMLP      & 76.25                   & 76.25                     & 76.25                     & 77.76    \\ \cline{2-6} 
\multicolumn{1}{|c|}{}                    & CWMLP      & 77.32                   & 77.22                     & 77.02                     & 78.7     \\ \hline
\multicolumn{1}{|c|}{\multirow{4}{*}{BK}} & \textbf{SPT\_CD}    & \textbf{100}            & \textbf{100}              & \textbf{100}              & \textbf{100}  \\ \cline{2-6} 
\multicolumn{1}{|c|}{}                    & CMLP       & 90                      & 90                        & 90                        & 90       \\ \cline{2-6} 
\multicolumn{1}{|c|}{}                    & SGMLP      & 100                     & 100                       & 100                       & 100      \\ \cline{2-6} 
\multicolumn{1}{|c|}{}                    & CWMLP      & 100                     & 100                       & 100                       & 100      \\ \hline
\end{tabular}
\end{table}

\begin{table*}
\centering
\caption{5-fold Cross-Validation accuracy results. 
Results obtained with our model (SPT\_CD) are compared with results published in~\cite{2018:Krakovna:interpretable} on the same dataset.}

\label{tab:krakovna}
\begin{tabular}{c|ccccccccccccc|c} 
\hline
 & \multicolumn{13}{|c|} {Krakovna \textit{et al.}~\cite{2018:Krakovna:interpretable} and La Grassa \textit{et al.}~\cite{LaGrassa}}  & \multicolumn{1}{c}{Our model} \\ \hline
Dataset                & Bart   & c5.0      & Cart            & Lasso        & LR            & NB       & RF            & SBFC      & SVM       & TAN       &  CD           & CD\_GP   &   N-ary    & \textbf{SPT\_CD}  \\ \hline
A                      & 71.6   & 66        & 63              & 65.6         & 52            & 69       & 71.8          & 72.2      & 72        & -         &  79.6         & 77.7     & 80.3       & \textbf{84}       \\ \hline
\end{tabular}
\end{table*}

\begin{table*}
\centering
\caption{5-fold Cross-Validation accuracy results on different datasets. 
Results obtained with our model (SPT\_CD) are compared with the results of different ensemble methods published.}
\label{tab:abpeykar}
\begin{tabular}{c|cccccccccccc|c} 
\hline
  & \multicolumn{12}{|c|} {Abpeykar \textit{et al.}~\cite{2019:Abpeykar} and La Grassa \textit{et al.}~\cite{LaGrassa}}      & \multicolumn{1}{c}{Our model}         \\ \hline
Dataset  & AdaB. & Bagg. & Dagg. & LogitB.        & Mod.            & Decor.  & Grad.   & Mt.B   & Stack.C      &  CD           & CD\_GP   &   N-ary   & \textbf{SPT\_CD}     \\ \hline
A        & 79.5  & 82.5  & 74.5  & 85.5           & \textbf{86.0}   & -       & 56.0    & 80.0   & 56.0         &  79.6         & 77.7     & 80.3      & 84.0                 \\ \hline
S        & 71.6  & 76.9  & 69.7  & 79.3           & 70.6            & 84.1    & 53.3    & 74.5   & 53.3         & 85.4          & 85.2     & \textbf{87.3}        & 84.1        \\ \hline
HL       & 50.4  & 50.2  & 50.4  & 50.4           & -               & -       & 50.4    & 50.4   & 50.4         & 58.1          & 57.9     &\textbf{ 61.1}      & 57.5        \\ \hline
G        & 88.9  & 75.0  & 82.2  & 89.4           & -               & 82.2    & 48.1    & 82.7   & 48.1         & 96.8          & -        & -         & \textbf{ 97.2 }                   \\ \hline
\end{tabular}
\end{table*}

\section{Conclusion and Future Work}
Multiclass problems can be solved by combining the decision boundary of many one-class classifiers trained on different classes. Finding the best partitions is a hard task but necessary to finalize the classification objective.
In this work, the methodology proposed has the aim to find out best partitions with ensemble method from the neighbourhood of an instance, based on subspace graph in training step and use this information as parameter in the objective function to improve final accuracy.
The presented results are competitive with classical classifiers and show a boost in many of them even when we use a few data to train the model. Against, it is computationally expensive to find the right combination of parameters and set up $\gamma$ value greater than 4.
A one-class classifier is trained using only one class and all rejected object are considered as outliers class.
Our method, as a combination of two one-class classifiers, does not consider outliers and all instances will be classified.
Future research efforts will focus on adapting the proposed approach to multi-class problems.
We will also apply the proposed approach to large scale datasets to evaluate the accuracy and time required.
Furthermore, we want to use deep feature extraction models and compare our work with well-known models typically used in computer vision tasks.
In order to reach these goals, we will focus on improving the computational complexity, modifying the algorithm and making it parallelizable.
Finally, we will evaluate the effectiveness of using different distance metrics and objective functions.

\bibliographystyle{IEEEtran}
\bibliography{bib}

\end{document}